# Rule Based Expert System for Cerebral Palsy Diagnosis

**\*Rajdeep Borgohain**
Department of Computer Science and Engineering, Dibrugarh University Institute of Engineering and Technology,
Dibrugarh, Assam
Email: rajdeepgohain@gmail.com
**Sugata Sanyal**
School of Technology and Computer Science, Tata Institute of Fundamental Research, Mumbai, India
Email: sanyals@gmail.com
**\*Corresponding Author**

------------------------------------------------------------------ABSTRACT----------------------------------------------------------------
*The use of Artificial Intelligence is finding prominence not only in core computer areas, but also in cross disciplinary areas including medical diagnosis. In this paper, we present a rule based Expert System used in diagnosis of Cerebral Palsy. The expert system takes user input and depending on the symptoms of the patient, diagnoses if the patient is suffering from Cerebral Palsy. The Expert System also classifies the Cerebral Palsy as mild, moderate or severe based on the presented symptoms.*
Keywords – Medical Expert System, Artificial Intelligence, Jess, Cerebral Palsy, Knowledge Base
-----------------------------------------------------------------------------------------------------------------------------------------------

## 1. INTRODUCTION[1]

With the advent of latest technologies in computer and information technology sector, the relationship between human and computer has reached a new level. The use of computer science is not only confined to the core computer areas like networks [7], network security [1, 4, 6], databases [2] etc. but also used in many cross disciplinary domains such as biology [12], chemistry [3], medical diagnosis [11] etc. In this paper, we present a rule based expert system for the diagnosis of Cerebral Palsy and classify it as mild, severe or moderate according to the symptoms presented by the users.

Expert System is software which has the ability to replicate the thinking and reasoning capacity of humans based on some facts and rules presented to it. The use of expert systems finds its place in diverse sectors like medical diagnosis, decision support systems, educational and tutorial software etc. The use of expert system in medical diagnosis dates back to the early 70's when MYCIN [13], an expert system for identifying bacteria causing diseases was developed at Stanford University. Since then, various expert systems like Internist – I, CADUCEUS etc. have been developed [14]. The goal of such an expert system is to aid the medical experts in making diagnosis of certain diseases or help the layman to diagnose the disease themselves.

In this paper we focus on Cerebral Palsy, which is a disorder that effects the body movement and posture due to some damage in the brain and spinal cord which had occurred during the development of the fetus, during birth of the baby or even during early childhood. According to a report by [5], Cerebral Palsy is the most commonly occurring motor function disorder among children. But many diseases which have similar symptoms like Pelizaeus-Merzbacher disease, Rett Syndrome, Charcot-Marie-Tooth disease can be mistaken for Cerebral Palsy without the advice of an expert medical professional [8]. In this case, the *Expert System for Cerebral Palsy Diagnosis* can aid the doctors and the caretakers of the patients in assessing if the disease is Cerebral Palsy or not.

In this paper, we look at the design and implementation of a rule based expert system which is used to diagnose and classify Cerebral Palsy as mild, moderate or severe by calculating the cumulative score of the weightage given to each symptoms.

The rest of the paper is organized in the following way: Section 2 gives an overview of Cerebral Palsy disease. Section 3 discusses the architecture of the proposed expert system. In Section 4, we look at the methodology for diagnosis. Section 5 discusses the evaluation of the system. Finally, we give a conclusion in Section 6.

## 2. CEREBRAL PALSY OVERVIEW

Cerebral Palsy is a non-progressive disease resulting from injury during fetal stage or early childhood which results in deficit of posture, movement, gait and tonus. Cerebral Palsy restricts the patient from performing full-fledged motor functions and limits the activities of the patient [15]. Statistics show that around the number of children suffering from Cerebral Palsy ranges from 1.5 to 4 per 1000 births. Cerebral Palsy not only affects the health condition but also affecting the economic conditions as medical costs for children with Cerebral Palsy were 10 times more than children without Cerebral Palsy or with other intellectual disabilities [5]. Depending on the symptoms of Cerebral Palsy, it can be classified as mild, moderate or severe.
The primary symptoms of Cerebral Palsy are:

1   Disease is non-progressive.
2   Spasticity in patients.
3   Disturbance in gait and mobility.
4   Abnormal Sensation.
5   Abnormal Perception
6   Impairment of Speech.

| | |
|---|---|
| 7 | Symptoms of mental retardation. |
| 8 | Involuntary and uncontrolled movements. |
| 9 | Disturbed sense of balance. |
| 10 | Stiff and Difficult movement. |
| 11 | Inability in controlling fine motor functions. |
| 12 | Awkward gait. |
| 13 | Joint Contractures. |
| 14 | Hearing Loss. |
| 15 | Loss of vision. |
| 16 | Bowel and bladder problems. |
| 17 | Dental Problems. |
| 18 | Drooling. |
| 19 | Postural Instability. |
| 20 | Symptoms appearing before 18 month of age. |

## 3. PROPOSED SYSTEM ARCHITECTURE

Our proposed system *Expert System for Cerebral Palsy Diagnosis* is a rule based expert system which has been developed using JESS, the Java Expert System Shell. The user of the expert system is first presented with a set of questionnaires to access the symptoms of the patient. Each symptom is given a certain weightage score so as to diagnose if the patient has the certain disease and classify it. The questionnaires are presented in simple English, which the user has to answer in affirmative or negative. According to the information provided by the user, the expert system makes use of the RETE algorithm to match the pattern facts with the rules. Once a certain rule is matched, the rule is fired and according to the rules stored in the knowledge base, the user is presented with a diagnosis.

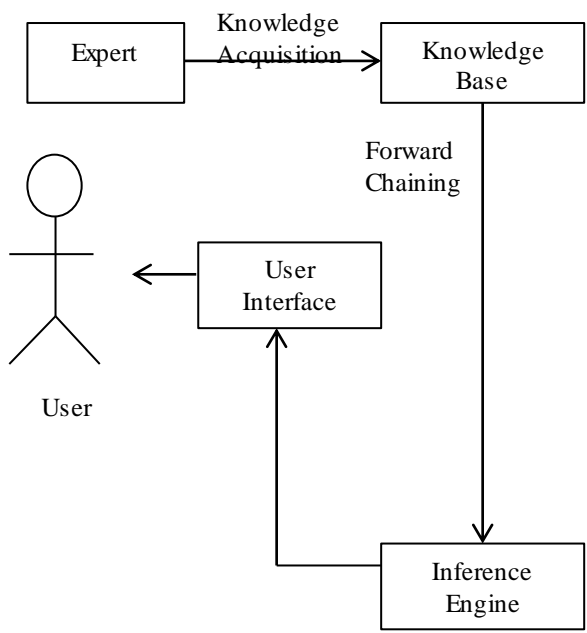

**Figure 1. Architecture of proposed system**

### 3.1.1 Knowledge Acquisition

The first and foremost work for building an Expert System is preparing a knowledge base for the system [16]. The primary source of information was interaction with doctors and postgraduate students of the Neurology department of the Assam Medical College. The second source of information was from the. The third source of acquisition of knowledge was from the internet.

### 3.1.2 Knowledge Representation

For knowledge representation, we used the Java Expert System shell (JESS) to represent facts and form the rules. First we present the questions before the user asking if the patient has suffered from the symptoms mentioned in table 1. The user either puts his answer as yes or no. We also take a global counter for the purpose of storing our cumulative weightage score. After each question is asked, we add the weightage score to the global counter if the answer is in affirmative and do not increment the counter if the answer to the particular questions is in negative. Finally, we get a cumulative score on the basis of which the disease is diagnosed.

Let us suppose, the patient is suffering from spasticity, then the rule in the JESS code is:

```
(defrule spastic
   (answer (ident spasticity) (text yes))
=>
 (bind *weightage* (+? *weightage* 5)))
```

Similarly for abnormal sensation we can have,

```
(defrule spastic
   (answer (ident abnormal-sensation) (text yes))
=>
 (bind *weightage* (+? *weightage* 1)))
```

Similarly, we make rules for every symptom. Finally, according to the cumulative score, the disease is classified accordingly.

The JESS code for classifying the disease is:

```
(defrule diagnosis
 ?p <- (result diagnosis-rule)
=>
 (printout t ? *weightage* crlf)

 (if (< ? *weightage* 16)) then
     (printout t "Symptoms show that you have no
     Cerebral Palsy." crlf))

 (if (>= ? *weightage* 16) (<= ? *weightage* 38)) then
     (printout t "Symptoms show that you have mild
     Cerebral Palsy." crlf))

  (if (>= ? *weightage* 39) (<= ? *weightage* 66)) then
        (printout t "Symptoms  show that you have
        moderate Cerebral Palsy." crlf))

 (if (>= ? *weightage* 66) (<= ? *weightage* 100)) then
     (printout t "Symptoms  show that you have severe
     Cerebral Palsy." crlf)))
```

### 3.1.3 *The RETE Algorithm*

The RETE algorithm is the core of the Java Expert System Shell for searching patterns in the rules. It is one of the most used algorithms for pattern searching. It highly speeds up the searching process by limiting the effort to re-compute the conflicts after a rule is fired [9]. The RETE algorithm is implemented as directed acyclic graphs which are used to match rules to facts [10].

## 4. METHODOLOGY FOR DIAGNOSIS

The diagnosis for Cerebral Palsy is done based on the answers given by the users to the questionnaires. For assessment of the disease, each symptom is given a certain weightage score [17] and based on this weightage score, a final cumulative weightage score is calculated. The disease is diagnosed according to this weightage score.

### 4.1 ALGORITHM FOR CUMULATIVE WEIGHTAGE SCORE

The algorithm for calculating the weightage score is:

1. The user of the system is presented with a questionnaire which contains queries relating to the patients symptoms. The questions are presented in simple English without any medical term for the convenience of the patients. The user of the system answers the questions in yes or no.

2. A list of symptoms with their corresponding weightage is presented in table 1. According to the input provided by the user, the corresponding weightage is assigned to each symptom. For an affirmative answer the weightage is assigned as given in the table while for a negative answer zero weightage is assigned.

3. Assess the cumulative score of the patient condition from the patient's response and normalize the score so that we get a percentage score according to the formula,

$$\text{Weightage Score} = \sum_{1}^{n} \frac{W_i}{C_i} \times 100$$

where, $W_i$ = *Weightage of ith symptom.*

$C_i$ = *Cumulative Score*

4. After tabulating the results, provide the information to the patient depending upon the cumulative score.

  I. No Cerebral Palsy (Below 16 %)
 II. Mild Cerebral Palsy (Between 16 % and 38%)
III. Moderate Cerebral Palsy (Between 39% and 66%)
 IV. Severe Cerebral Palsy (Above 66%)

| Rule # | Classical Symptoms | Weightage |
|---|---|---|
| 1. | Non Progressive | 5 |
| 2. | Spasticity | 5 |
| 3. | Disturbance in gait and mobility | 5 |
| 4. | Abnormal sensation | 1 |
| 5. | Abnormal perception | 1 |
| 6. | Impairment of Speech | 1 |
| 7. | Mental Retardation | 3 |
| 8. | Involuntary and Uncontrolled Movement | 2 |
| 9. | Disturbed Sense of Balance | 2 |
| 10. | Stiff and difficult movement | 5 |
| 11. | Inability to control fine motor function | 1 |
| 12. | Awkward Gait | 1 |
| 13. | Joint Contractures | 5 |
| 14. | Hearing Loss | 1 |
| 15. | Vision Loss | 1 |
| 16. | Bowel Bladder Problems | 2 |
| 17. | Dental Problems | 1 |
| 18. | Drooling | 2 |
| 19. | Postural Instability | 2 |
| 20. | Symptoms appear before 18 months | 5 |

**Table 1. Weightage of Symptoms**

## 5. EVALUATION OF THE SYSTEM

The system was evaluated by testing against a few proved of Cerebral Palsy as test cases. The system showed accurate results when the specific symptoms of the test cases were given as user input. Moreover, a number of parents having patients already diagnosed with Cerebral Palsy were asked for their input. The system showed accurate result and diagnosed the patients as Cerebral Palsy patients and classified their diseases accordingly.

## 6. CONCLUSION

In this paper we have discussed the design and implementation of a rule based *Expert System for Cerebral Palsy Diagnosis.* The expert system helps to diagnose Cerebral Palsy and classify it as mild, moderate or severe. In the implementation, we have taken the most classical symptoms of Cerebral Palsy and given a weightage to each of the symptom and according to the feedback given by the user. The expert system can go a great deal in supporting the decision making process of medical professionals and also help parents having children with Cerebral Palsy to assess their children and to take appropriate measures to manage the disease.